\documentclass[twoside,11pt]{article}

\usepackage{jmlr2e}
\usepackage{algorithm}
\usepackage{algorithmic}
\usepackage{multirow}
\usepackage{amsmath}
\usepackage{xspace} 
\usepackage{colortbl}
\usepackage{caption}
\usepackage{subcaption}
\usepackage{bm, graphicx}

 \newcommand{\w}{{\boldsymbol \theta}}
 \newcommand{\tr}{\top}     
 \newcommand{\x}{{\boldsymbol \phi}}
 \newcommand{\bs}{\boldsymbol}

\ShortHeadings{An Empirical Evaluation of True Online TD($\lambda$)}{}
\firstpageno{1}
\title{An Empirical Evaluation of True Online TD($\lambda$)}

\author{\name{Harm van Seijen} \email{harm.vanseijen@ualberta.ca}\\
       \name{A. Rupam Mahmood} \email{ashique@ualberta.ca}\\
       \name{Patrick M. Pilarski} \email{patrick.pilarski@ualberta.ca}\\
       \name Richard S. Sutton \email sutton@cs.ualberta.ca \\
       \addr Department of Computing Science\\
       University of Alberta\\
       T6G 2E8, Canada}

\begin{document}
\maketitle

\begin{abstract}

The true online TD($\lambda$) algorithm has recently been proposed \citep{vanseijen:icml14} as a universal replacement for the popular TD($\lambda$) algorithm, in temporal-difference learning and reinforcement learning.  True online TD($\lambda$) has better theoretical properties than conventional TD($\lambda$), and the expectation is that it also results in faster learning. In this paper, we put this hypothesis to the test. Specifically, we compare the performance of true online TD($\lambda$) with that of TD($\lambda$) on challenging examples, random Markov reward processes, and a real-world myoelectric prosthetic arm. We use linear function approximation with tabular, binary, and non-binary features. We assess the algorithms along three dimensions: computational cost, learning speed, and ease of use.
Our results confirm the strength of true online TD($\lambda$): 1) for sparse feature vectors, the computational overhead with respect to TD($\lambda$) is minimal; for non-sparse features the computation time is at most twice that of TD($\lambda$), 2) across all domains/representations the learning speed of true online TD($\lambda$) is often better, but never worse than that of TD($\lambda$), and 3) true online TD($\lambda$) is easier to use, because it does not require choosing between trace types, and it is generally more stable with respect to the step-size. Overall, our results suggest that true online TD($\lambda$) should be the first choice when looking for an efficient, general-purpose TD method.

\end{abstract}


\section{Introduction}

Temporal-difference (TD) learning  \citep{sutton:ml88} is a core learning technique in modern reinforcement learning \citep{kaelbling:jair96, sutton:book98, szepesvari:book10}. One of the main challenges in reinforcement learning is to make predictions, in an initially unknown environment, about the (discounted) sum of future rewards, the return, based on currently observed feature values and a certain behaviour policy. With TD learning it is possible to learn good estimates of the expected return quickly by bootstrapping from other expected-return estimates. TD($\lambda$) \citep{sutton:ml88} is a popular TD algorithm that combines basic TD learning with eligibility traces to further speed learning. 

The ability of TD($\lambda$) to speed learning is explained by its forward view, which states that the estimate at each time step is moved toward an update target known as the $\lambda$-return, where the $\lambda$-parameter determines the trade-off between bias and variance of the update target. This trade-off has a large influence on the speed of learning and its optimal setting varies from domain to domain. The ability to improve this trade-off by adjusting the value of $\lambda$ enables eligibility traces to improve the learning speed.

True online TD($\lambda$) \citep{vanseijen:icml14} is a recently proposed variation of TD($\lambda$) with better theoretical properties. Specifically, it maintains an exact equivalence to the forward view at all times. In contrast, TD($\lambda$) accurately approximates the forward view only for appropriately small step-sizes. Hence, it can be expected that true online TD($\lambda$) can do a better job in improving the learning speed. Initial experiments suggest that this is indeed the case  \citep{vanseijen:icml14}. However, no significant empirical study has been performed so far. In this paper, we empirically compare true online TD($\lambda$) with TD($\lambda$) on a wide variety of domains. 

\section{Markov Reward Processes}

We focus in this paper on discrete-time \emph{Markov reward processes} (MRPs), which can be described as 4-tuples of the form $\langle \mathcal{S}, p, r, \gamma \rangle$, consisting of $\mathcal{S}$, the set of all states; $p(s'|s)$, the transition probability function, giving for each state $s \in \mathcal{S}$ the probability of a transition to state $s' \in \mathcal{S}$ at the next step;  $r(s,s')$, the reward function, 
giving the expected reward after a transition from $s$ to $s'$. $\gamma$ is the discount factor, specifying how future rewards are weighted with respect to the immediate reward. An MRP can contain \emph{terminal states}, dividing the sequence of state transitions into \emph{episodes}. When a terminal state is reached the current episode ends and the state is reset to the initial state. 
The \emph{return} at time step $t$ is the discounted sum of rewards observed after time step $t$:
$$G_t =  \sum_{i=1}^\infty  \gamma^{i-1} R_{t+i}\thinspace,$$
where $R_k$ is the reward received at time $k$.
For an episodic MRP, the return is the discounted sum of rewards up to the time step that the terminal state is reached.

We are interested in learning the value-function $v$ of an MRP, which maps each state $s \in \mathcal{S}$ to the expected value of the return:
$$v(s) = \mathbb{E}\{ G_t \,|\, S_t = s\}\thinspace.$$
In the general case, the learner does not have access to $s$ directly, but can only observe a feature vector $\x(s) \in \mathbb{R}^n$. We estimate the value function using linear function approximation, in which case the value of a state is the inner product between a weight vector $\w$ and a feature vector $\x$. In this case, the value of state $s$ is approximated by:
$$ \hat v (s, \w) = \w^\tr \x(s)  = \sum_{i=1}^n \theta_i\, \phi_i(s)\,.$$
As a shorthand, we will indicate $\x({S_t})$, the feature vector of the state visited at time step $t$, by $\x_t$. 

\section{Algorithms}

This section presents the two methods that we compare: conventional TD($\lambda$) and true online TD($\lambda$).

\subsection{Conventional TD($\lambda$)}

The conventional TD($\lambda$) algorithm is defined by the following update equations:
\begin{eqnarray}
\delta_t &=& R_{t+1} + \gamma \w_t^\tr \x_{t+1}   - \w_{t}^\tr \x_{t} \label{eq:delta_update}\\
{\bs e}_t &=& \gamma\lambda {\bs e}_{t-1} +  \x_t  \label{eq:trace_update}\\
\w_{t+1} &=&  \w_t + \alpha \delta_t\,{\bs e}_{t}  \label{eq:weight_update}
\end{eqnarray}
for $t \geq 0$, and with ${\bs e}_{-1} = {\bs 0}$. The scalar $\delta_t$ is called the \emph{TD error}. The vector ${\bs e}_t$ is called the \emph{eligibility-trace} vector, and the parameter $\lambda \in [0, 1]$ is called the \emph{trace-decay} parameter.

TD($\lambda$) can be very sensitive with respect to the $\alpha$ and $\lambda$ parameters. Especially, a large value of $\lambda$ combined with a large value of $\alpha$ can easily cause divergence, even on simple tasks with bounded rewards. For this reason, a variant of TD($\lambda$) is often used that is more robust with respect to these parameters. This variant, which assumes binary features, uses a different trace-update equation:
 \begin{displaymath}
e_{t, i}= \begin{cases} \gamma\lambda  e_{t-1, i} &\mbox{if } \phi_i (S_t) = 0\\
1 & \mbox{if }  \phi_i (S_t)= 1 \end{cases} 
\qquad\mbox{ for all } i\thinspace.
\end{displaymath}
When TD($\lambda$) uses this equation to update its elegibility-trace vector, it is said to use \emph{replacing traces}; in contrast, the default implementation based on trace update (\ref{eq:trace_update}) is said to use \emph{accumulating traces}. 
In this paper, we will indicate these implemenations by `replace TD($\lambda$)' and `accumulate TD($\lambda$)', respectively. In our experiments, we compare against both versions.

\subsection{True Online TD($\lambda$)}

The true online TD($\lambda$) update equations are:
\begin{eqnarray}
\delta_t &=& R_{t+1} + \gamma \w_t^\tr \x_{t+1}   - \w_{t}^\tr \x_{t} \label{eq:to_delta_update}\\
{\bs e}_t &=& \gamma\lambda {\bs e}_{t-1} +  \x_t -  \alpha \gamma \lambda [{\bs e}_{t-1}^\tr \,\x_t] \,\x_t \label{eq:to_trace_update}\\
\w_{t+1} &=&  \w_t + \alpha \delta_t\,{\bs e}_{t} + \alpha [ \w_{t}^\tr \x_{t} - \w_{t - 1}^\tr \x_{t} ] [ {\bs e}_t -  \x_t ]\quad \label{eq:to_weight_update}
\end{eqnarray}
for $t \geq 0$, and with ${\bs e}_{-1} = {\bs 0}$. Compared to accumulate TD($\lambda$), both the trace update and the weight update have an additional term. We call a trace updated in this way a \emph{dutch trace};  we call the term $\alpha [ \w_{t}^\tr \x_{t} - \w_{t - 1}^\tr \x_{t} ] [ {\bs e}_t -  \x_t ]$ the \emph{TD-error time-step correction}, or simply the $\delta$-correction.
For $\lambda = 0$, ${\bs e_t}$ reduces to $\x_t$, in which case the $\delta$-correction is 0. Hence, for $\lambda = 0$, true online TD($\lambda$) reduces to the regular TD(0) method. 

Algorithm \ref{al:true TD(lambda)} shows pseudocode that implements true online TD($\lambda$).\footnote{In the pseudocode $\hat v_{old}$ is initialized to 0, but any (finite) value would do, because for the first update of an episode, ${\bs e}_t$ is equal to $\x_t$, and hence the $\delta$-correction is equal to zero, regardless of the value of $\hat v_{old}$.} In order to discuss its computational cost, let $n$ be the total number of features, and $m$ the number of features with a non-zero value. Then, the number of basic operations (addition and multiplication) per time step for conventional TD($\lambda$) is $3n + 5m$. True online TD($\lambda$) takes another $6m$, resulting in $3n + 11m$ operations in total.\footnote{Note that computing and adding the vector $\alpha \gamma\lambda  ({\bs e}^\tr \x )\,\x$ requires only $4m$ operations.} Hence, if sparse feature vectors are used (that is, if $m << n$) the computational overhead of true online TD($\lambda$) is minimal. If non-sparse feature vectors are used (that is, $m = n$) TD($\lambda$) and true online TD($\lambda$) require $8n$ and $14n$, respectively. So in this case, true online TD($\lambda$) is roughly twice as expensive as conventional TD($\lambda$).


\begin{algorithm}[thb]
\begin{algorithmic}[0]
\STATE {\bf INPUT: $\alpha, \lambda, \gamma, \w_{init}$}
\STATE $\w \leftarrow \w_{init}, \,\, {\hat v}_{old} \leftarrow 0$
\STATE Loop (over episodes):
\STATE \qquad obtain initial $\x$
\STATE \qquad${\bs e} \leftarrow {\bs 0}$
\STATE \qquad While terminal state has not been reached, do:
\STATE \qquad\qquad obtain next feature vector $\x'$ and reward $R$
\STATE \qquad\qquad $ \hat v \leftarrow \w^\tr\x$
\STATE \qquad\qquad $\hat v' \leftarrow \w^\tr\x'$
\STATE \qquad\qquad $\delta \leftarrow R + \gamma\, \hat v' - \hat v$
\STATE \qquad\qquad $ {\bs e} \leftarrow  \gamma\lambda {\bs e}  + \x - \alpha \gamma\lambda  ({\bs e}^\tr \x )\,\x$
\STATE \qquad\qquad $\w \leftarrow  \w + \alpha  (\delta + \hat v - \hat v_{old})\,  {\bs e} - \alpha (\hat v - \hat v_{old})\x$
\STATE  \qquad\qquad $\hat v_{old} \leftarrow \hat v'$
\STATE \qquad\qquad $\x \leftarrow \x' $
\caption{true online TD($\lambda$)}
\label{al:true TD(lambda)}
\end{algorithmic}
\end{algorithm}

\section{Empirical Comparisons}

In this section, we compare the performance of true online TD($\lambda$) with that of accumulate TD($\lambda$) and replace TD($\lambda$). First, we compare the behaviour on two challenging examples, then on random MRPs, and finally on a real-world data set. For the experiments on the random MRPs and the real-world data set, all methods use the same sample sequence (and they start from the same initial values). Hence, a lower error corresponds with a higher learning speed.

\subsection{Challenging Examples}

\begin{figure}[thb]
\begin{center}
\includegraphics[width=13cm]{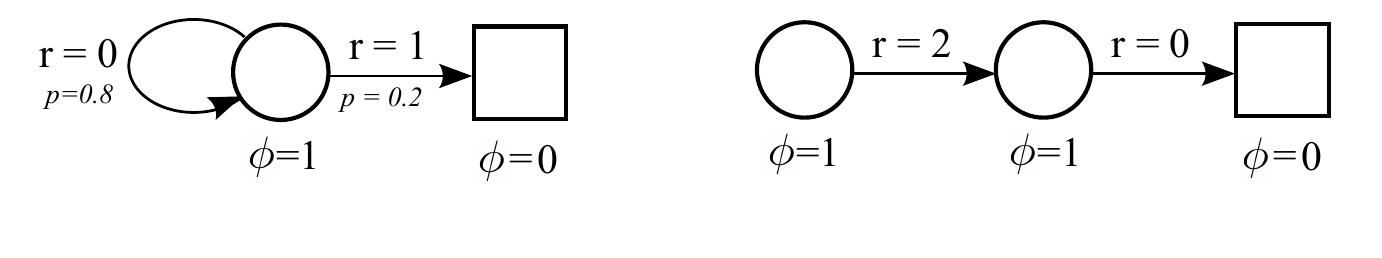}
\caption{{\it Left: } One-state example. {\it Right: } Two-state example. Circles indicate states; squares are terminal states; arrows indicate state transitions ($p$ is the transition probability). In both examples, the state-space is represented by a single, binary feature $\phi$.}
\label{fig:counter_examples}
\end{center}
\end{figure}

For our first experiments, we designed two small examples that are challenging for either accumulate TD($\lambda$) or replace TD($\lambda$) to check if true online TD($\lambda$) can deal well with such problems (see Figure \ref{fig:counter_examples} ). The first one, a one-state example, is challenging for accumulate TD($\lambda$), while the second one, a two-state example, is challenging for replace TD($\lambda$). The challenging part of the one-state example is that the same feature is revisited frequently within the same episode; the challenging part of the two-state example is that the value function cannot be represented exactly (because the two states have different values, but are represented in the same way).

\begin{figure}[!h]
\begin{center}
\includegraphics[width=6cm]{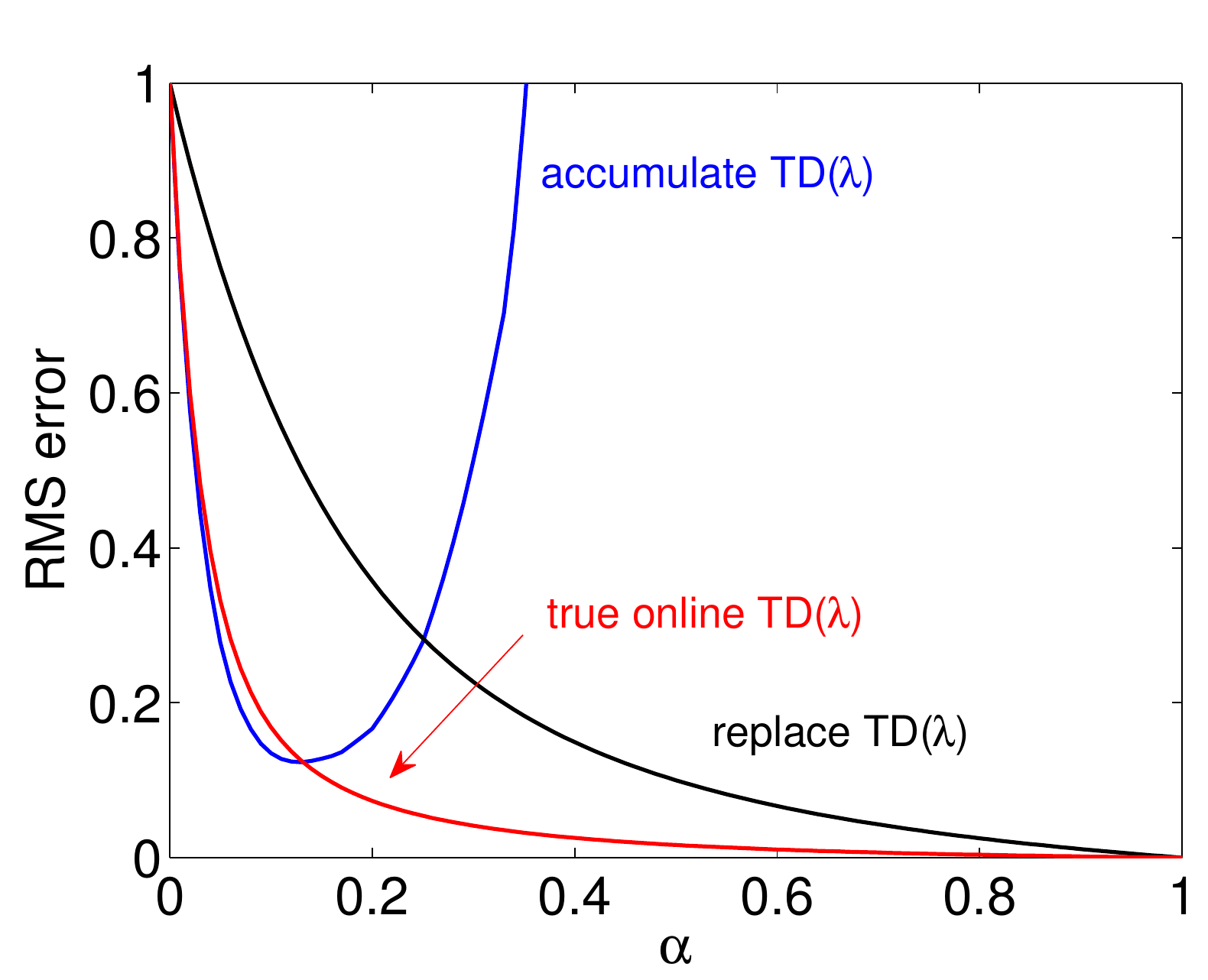}
\hspace{1cm}
\includegraphics[width=6cm]{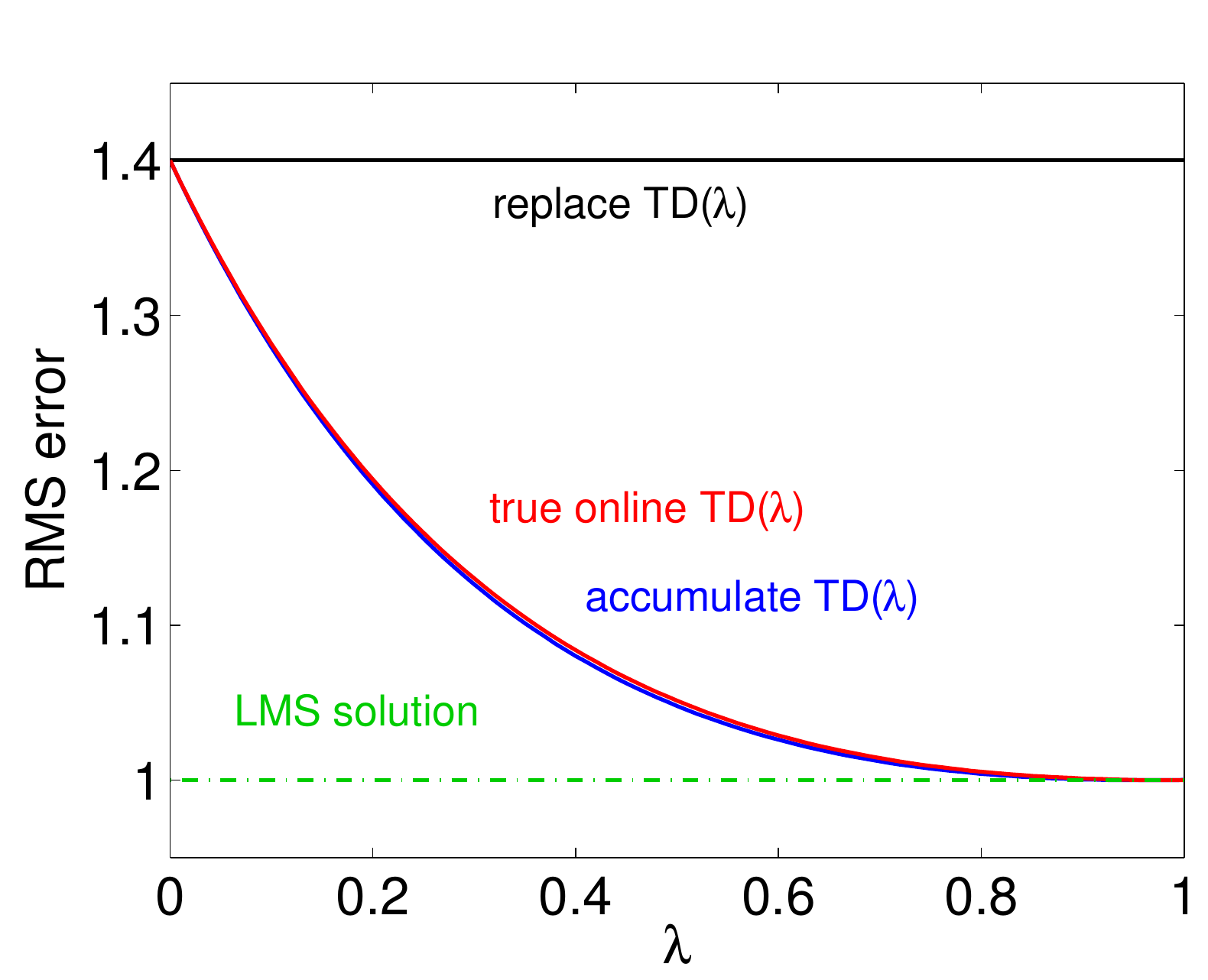}
\caption{{\it Left: }  RMS error at the end of an episode, averaged over the first 10 episodes, on the one-state example (using $\lambda = 1$). {\it Right: } RMS error after approximate convergence on the two-state example (using $\alpha = 0.01$). We considered values to be converged if the error changed less than 1\% over the last 100 time steps. }
\label{fig:performance_counter_examples}
\end{center}
\end{figure}

The left graph of Figure \ref{fig:performance_counter_examples} shows the early learning performance on the one-state example at $\lambda = 1$ and for different step-size values. 
Replace  TD($\lambda$) and true online TD($\lambda$) do very well on this task. For $\alpha = 1$, the state-value converges after a single episode already, resulting in an average RMS error of zero. This is not surprising, given that the return has zero variance. However, the error for accumulate TD($\lambda$) diverges at these settings, and even with optimized step-size, the RMS error does not reach zero. 

The right graph of Figure \ref{fig:performance_counter_examples} shows the RMS error after approximate convergence for different $\lambda$ values on the two-state example. Also the RMS error for the least mean squares (LMS) solution is shown. For accumulate TD($\lambda$) the error is equal to the LMS solution for $\lambda = 1$, but gets worse for smaller $\lambda$. This corresponds with the theory, which states that the fixed point of accumulate TD($\lambda$) equals the LMS solution for $\lambda = 1$, but is different from the LMS solution for  $\lambda < 1$  \citep{dayan:ml92}. True online TD($\lambda$) has the same behaviour. Surprisingly, for replace TD($\lambda$) the value of $\lambda$ has no effect. This task illustrates the main weakness of replace TD($\lambda$): while it avoids the divergence issues of accumulate TD($\lambda$), it is an overly conservative approach. It avoids divergence by resetting a trace each time a feature is revisited, which reduces the overall effect of a trace. In the extreme case, the effect can be removed completely, which is what happens here.

Overall, these tasks show that while accumulate TD($\lambda$) and replace TD($\lambda$) each have their weakness, true online TD($\lambda$) does not suffer from these weaknesses. Of course, the problems we constructed here are extreme examples. In practise, tasks will be not be so one-sided, but they can have properties from both examples.

\subsection{Random MRPs}

For our second series of experiments we used randomly constructed MRPs.\footnote{The process we used to construct these MRPs is based on the process used by Bhatnagar, Sutton, Ghavamzadeh and Lee (2009).}  We represent a random MRP as a 3-tuple $(k, b, \sigma)$, consisting of $k$, the number of states; $b$, the branching factor (that is, the number of possible next states per transition); and $\sigma$, the standard deviation of the reward  (the expected value is drawn at random from a normal distribution with zero mean). We compared the performance on three different MRPs: one with a small number of states, $(10, 3, 0.1)$, one with a larger number of states, $(100, 10, 0.1)$, and one with a low branching factor and no stochasticity in the reward, $(100, 3, 0)$. We evaluated each MRP using three different representations: one with tabular features, one with binary features and one with non-binary, normalized features (normalized such that the length of the feature vector is always 1). For more details on the representations, see appendix A. For each domain/representation/method combination we performed a scan over $\alpha$ and $\lambda$ values to determine the best performance for each combination. As performance metric we used the mean-squared error (MSE) with respect to the LMS solution during early learning (for $k=10$, we averaged over the first $100$ time steps; for $k=100$, we averaged over the first $1000$ time steps). We normalized this error by dividing it by the MSE error obtained for $\lambda = 0$ for the relevant domain/representation combination (all three TD($\lambda$) methods reduce to the same algorithm for $\lambda = 0$). In addition, we averaged over 50 independent runs. For each domain/representation/method we used the same (randomly generated) sample sequences.\footnote{The code for the MRP experiments is published online at: https://github.com/armahmood/totd-rndmdp-experiments}

\begin{figure}[thb]
\begin{center}
\hspace{-2cm}
\includegraphics[width=17cm]{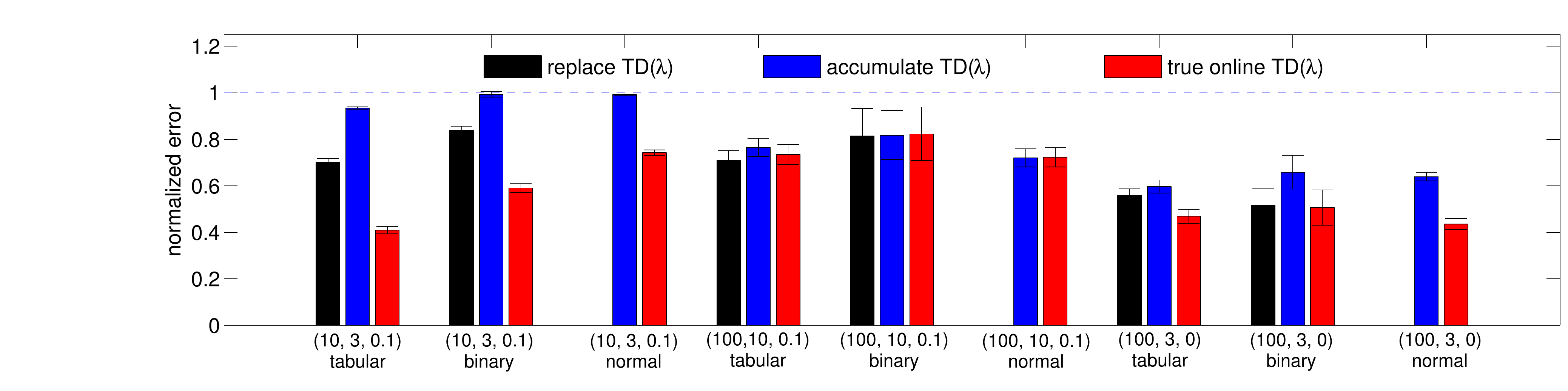}
\caption{Normalized MSE error at the best parameter settings for all MRP experiments. The error is normalized by dividing it by the MSE error obtained for $\lambda = 0$.}
\label{fig:results summary}
\end{center}
\end{figure}

Figure \ref{fig:results summary} shows the results of the comparisons for each domain/representation combination. Because  $\lambda = 0$ lies in the parameter range that is being optimized over, the normalized error can never be higher than 1. If for a method/domain the normalized error is equal to 1, this means that setting $\lambda$ higher than 0 either has no effect, or that the error gets worse. In either case, eligibility traces are not effective for that domain/representation/method combination.

The results confirm the strength of true online TD($\lambda$). The optimal performance of true online TD($\lambda$) is, on all domains and for all representations, at least as good as the optimal performance of replace TD($\lambda$) and of accumulate TD($\lambda$). Specifically, true online TD($\lambda$) outperforms conventional TD($\lambda$) on 5 of the 9 domains/representations considered. In the next subsection, we compare the methods using real-world data.

\subsection{Predicting Signals from a Myoelectric Prosthetic Arm}

\begin{figure}[tbp]
\begin{center}
\includegraphics[width=1.8in]{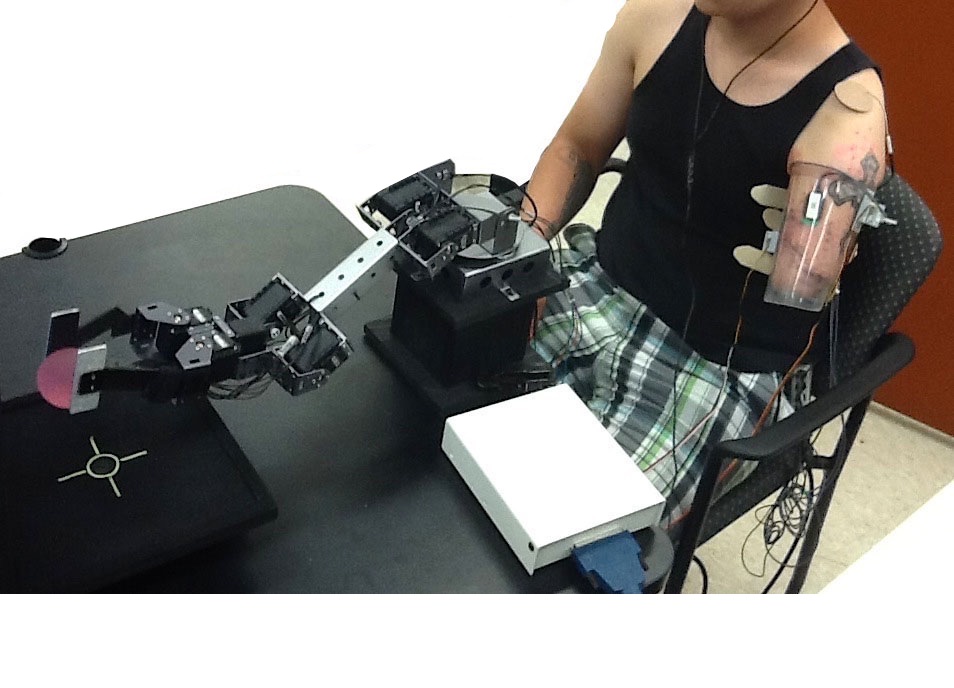}
\includegraphics[width=1.8in]{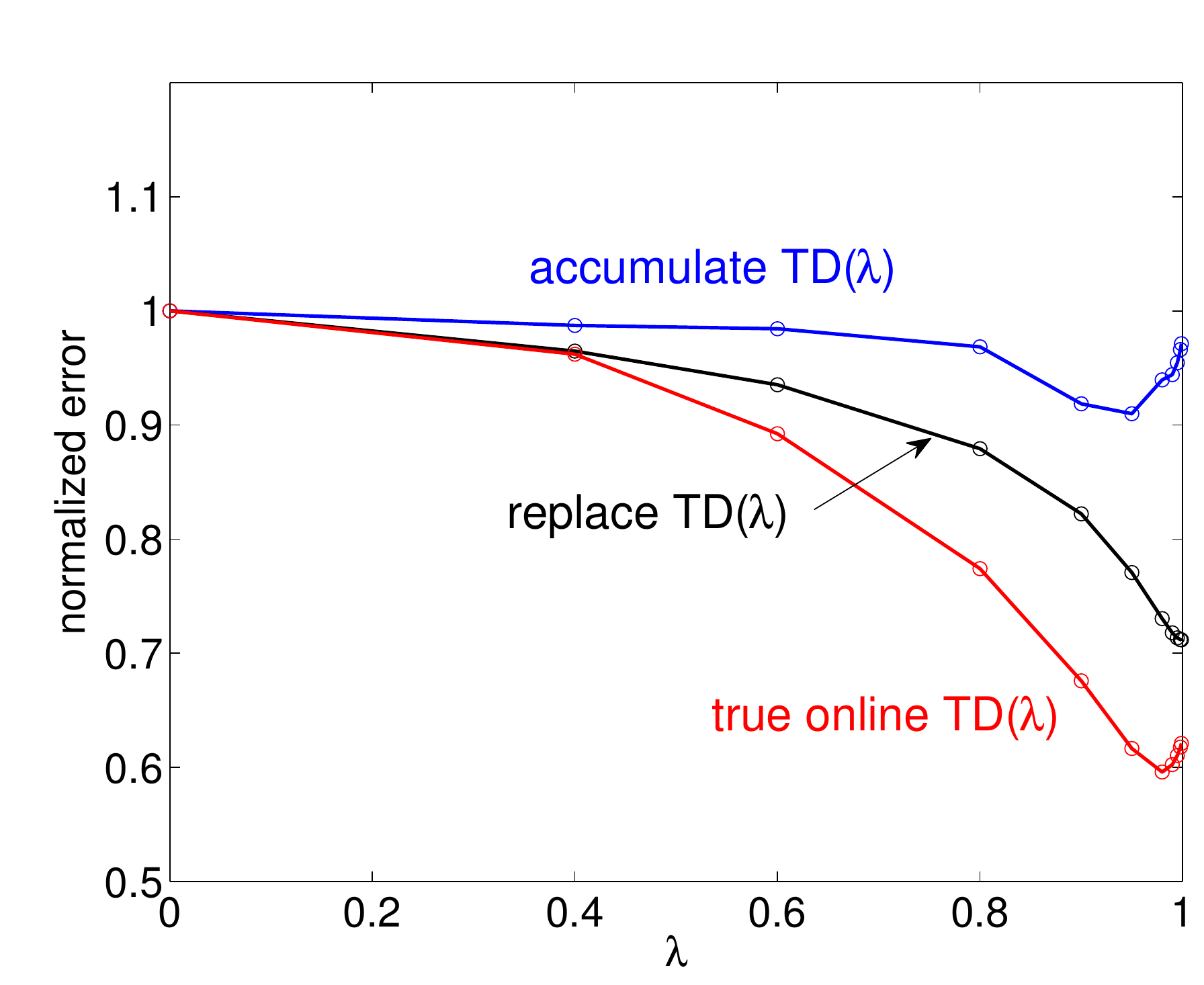}
\includegraphics[width=1.8in]{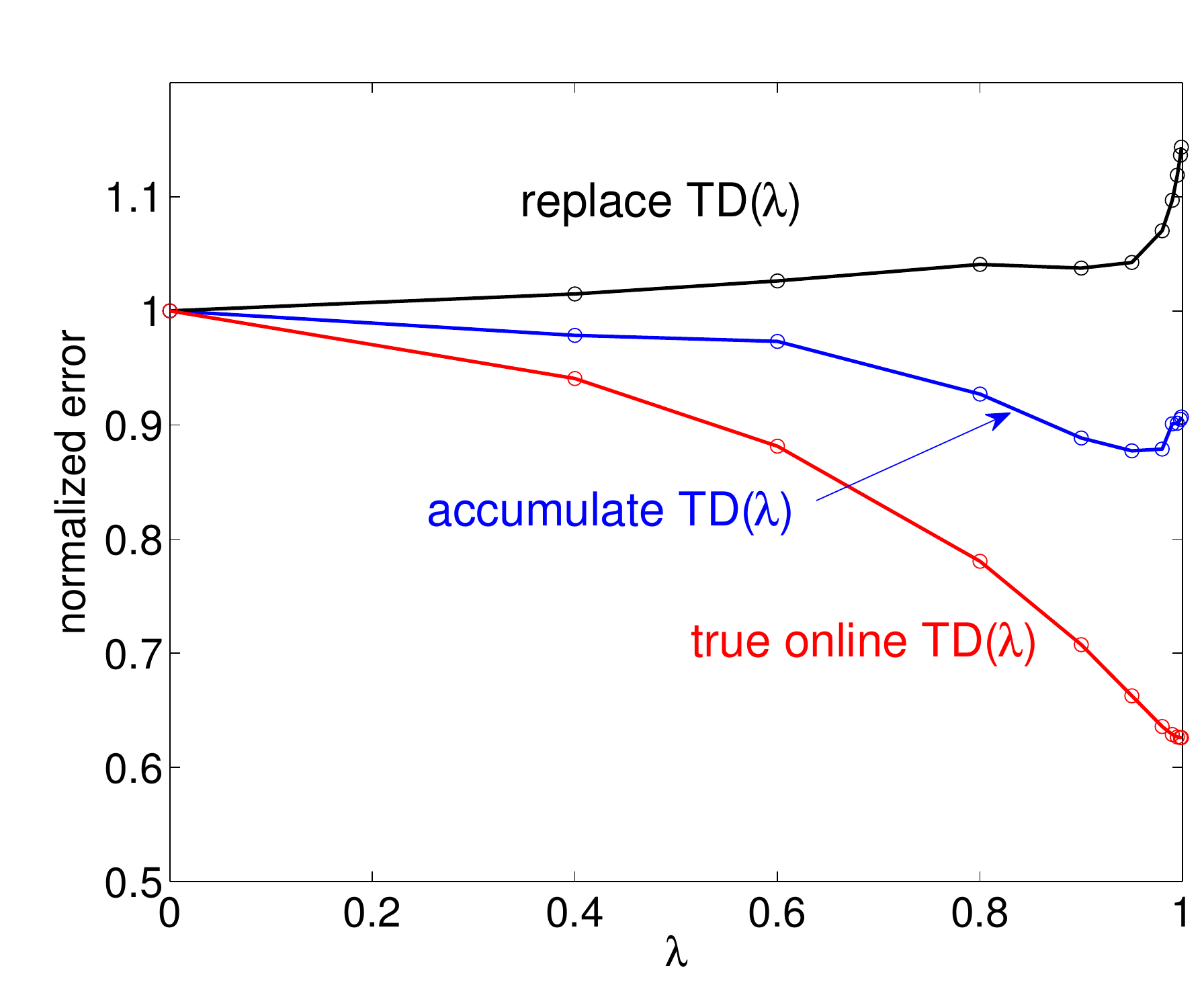}
\caption{{\it Left: } picture of experimental setup. {\it Middle: } normalized error of the predictions for different $\lambda$ at the best $\alpha$ value, for the force predictions. {\it Right: } same, but for angle predictions.}
\label{fig:myoelectricresults}
\end{center}
\end{figure}

In this experiment, we compare the performance of true online TD($\lambda$) and conventional TD($\lambda$) on a real-world data-set consisting of sensorimotor signals measured during the human control of an electromechanical robot arm. The source of the data is a series of manipulation tasks performed by a participant with an amputation, as presented by \cite{pilarski:ra13}. In this study, an amputee participant used signals recorded from the muscles of their residual limb to control a robot arm with multiple degrees-of-freedom (Figure \ref{fig:myoelectricresults}, left). Interactions of this kind are known as {\em myoelectric control} \citep[c.f.,][]{parker:jek06}. 

For consistency and comparison of results, we used the same source data and prediction learning architecture as published in \cite{pilarski:ra13}. In total, two signals are predicted: grip force and motor angle signals from the robot's hand. 
Specifically, the target for the prediction is a discounted sum of each signal over time, similar to return predictions \citep[c.f., general value functions and nexting;][]{sutton:aamas11, modayil:ab14}.
Where possible, we used the same implementation and code base as \cite{pilarski:ra13}. Data for this experiment consisted of 58,000 time steps of recorded sensorimotor information, sampled at 40 Hz (i.e., approximately 25 minutes of experimental data). The state space consisted of a tile-coded representation of the robot gripper's position, velocity, recorded gripping force, and two muscle contraction signals from the human user.  A standard implementation of tile-coding was used, with ten bins per signal, eight overlapping tilings, and a single active bias unit. This results in a state space with 800,001 features, 9 of which were active at any given time.  Hashing was used to reduce this space down to a vector of 200,000 features that are then presented to the learning system.  All signals were normalized between 0 and 1 before being provided to the function approximation routine. The discount factor for predictions of both force and angle was  $\gamma=0.97$, as in the results presented by \cite{pilarski:ra13}. Parameter sweeps over  $\lambda$ and $\alpha$ are conducted for all three methods.
The performance metric is the mean absolute return error over all 58,000 time steps of learning, normalized by dividing by the error for $\lambda = 0$. 

Figure  \ref{fig:myoelectricresults} shows the performance for the angle as well as the force predictions. The relative performance of replace TD($\lambda$) and accumulate TD($\lambda$) depends on the predictive question being asked. For predicting the robot's grip force signal---a signal with small magnitude and rapid changes---replace TD($\lambda$) is better  than accumulate TD($\lambda$) at all non-zero $\lambda$ values. However, for predicting the robot's hand actuator position, a smoothly changing signal that varies between a range of $\sim$300-500, accumulate TD($\lambda$)  dominates replace TD($\lambda$) over all non-zero $\lambda$ values. True online TD dominates both methods for all non-zero $\lambda$ values on both prediction tasks (force and angle).

\section{Conclusions}

We have compared true online TD($\lambda$) to conventional TD($\lambda$) along three broad dimensions: computational cost, learning speed, and ease of use. 
In terms of computational cost, TD($\lambda$) has a slight advantage.
In the worst case, true online TD($\lambda$) is  twice as expensive. In the typical case of sparse features, it is only fractionally more expensive than TD($\lambda$).
Memory requirements are the same for both methods.
In terms of learning speed, in our experiments true online TD($\lambda$) was usually better and never worse than TD($\lambda$).
Specifically, true online TD($\lambda$) substantially outperformed TD($\lambda$)  on 5 out of the 9 MRPs and in both myoelectric-arm experiments. 
Finally, in terms of ease of use, we conclude that true online TD($\lambda$) has a clear advantage. 
The first difficulty with conventional TD($\lambda$) is that typically one must choose between its two types of traces, whereas with true online TD($\lambda$) no such choice has to be made.
A second difficulty for accumulate TD($\lambda$) is that its performance can be very sensitive to the step-size parameter (e.g., see Figure \ref{fig:performance_counter_examples}, left), making it hard to find an acceptable value.
Overall, our results suggest that true online TD($\lambda$) should be the first choice when looking for an efficient, general-purpose TD method.

\appendix

\section{Details of MRP Experiments}

Let $k$ be the number of states in a domain.
For the tabular representation each state is represented with a unique standard-basis vector of $k$ dimensions. The binary representation is constructed by first assigning indices, from 1 to $k$, to all states. Then, the binary encoding of the index of a state is used as a feature vector to represent that state. The length of a feature vector is determined by the total number of states: for $k = 10$, the length is $4$;  for $k= 100$, the length is $7$. As an example, for $k= 10$ the feature vectors of states $1$, $2$ and $3$ are $(0,0,0,1), (0,0,1,0)$ and $(0,0,1,1)$, respectively. Finally, for the non-binary, normal representation each state is mapped to a 5-dimensional feature vector, with the value of each feature drawn from a normal distribution with zero mean and unit variance. After all the feature values for a state are drawn, they are normalized such that the feature vector has unit length. Once generated, the feature vectors are kept fixed for each state. 


\end{document}